\documentclass[letterpaper]{article} 
\usepackage{aaai25}  
\usepackage{times}  
\usepackage{helvet}  
\usepackage{courier}  
\usepackage[hyphens]{url}  
\usepackage{graphicx} 
\usepackage{natbib}  
\usepackage{caption} 
\usepackage{algorithm}
\usepackage{algorithmic}
\usepackage{amsmath}
\usepackage{amssymb}
\usepackage{textcomp}
\usepackage{url}
\usepackage{verbatim}
\usepackage{graphicx}
\usepackage{cite}
\usepackage{multirow}
\usepackage{colortbl}
\usepackage{pifont}
\usepackage{booktabs}
\usepackage{color}
\usepackage{colortbl}
\usepackage[table]{xcolor}
\usepackage{newfloat}
\usepackage{listings}
\DeclareCaptionStyle{ruled}{labelfont=normalfont,labelsep=colon,strut=off} 
\floatstyle{ruled}
\newfloat{listing}{tb}{lst}{}
\floatname{listing}{Listing}
\title{Hierarchical Alignment-enhanced Adaptive Grounding Network for Generalized Referring Expression Comprehension}
\author{
    Yaxian Wang\textsuperscript{\rm 1,2},
    Henghui Ding\textsuperscript{\rm 3}\thanks{Corresponding authors. henghui.ding@gmail.com},
    Shuting He\textsuperscript{\rm 4},
    Xudong Jiang\textsuperscript{\rm 5},
    Bifan Wei\textsuperscript{\rm  6,7$\ast$},
    Jun Liu\textsuperscript{\rm 1,2}
}

\affiliations{
    \textsuperscript{\rm 1}School of Computer Science and Technology, Xi’an Jiaotong University, China\\
       \textsuperscript{\rm 2}Ministry of Education Key Laboratory of Intelligent Networks and Network Security, Xi’an Jiaotong University, China
   \textsuperscript{\rm 3}Institute of Big Data, Fudan University, China\\ 
      \textsuperscript{\rm 4}Shanghai University of Finance and Economics, China\\ \textsuperscript{\rm 5}Nanyang Technological University, Singapore\\
       \textsuperscript{\rm 6}School of Continuing Education, Xi’an Jiaotong University, China\\
       \textsuperscript{\rm 7}Shaanxi Province Key Laboratory of Big Data Knowledge Engineering, Xi’an Jiaotong University, China\\
    wyx1566@stu.xjtu.edu.cn,  henghui.ding@gmail.com,
    shuting.he@sufe.edu.cn,
\{weibifan,liukeen\}@xjtu.edu.cn
       
}

\usepackage{bibentry}

\begin{document}

\maketitle

\begin{abstract}
In this work, we address the challenging task of Generalized Referring Expression Comprehension (GREC). Compared to the classic Referring Expression Comprehension (REC) that focuses on single-target expressions, GREC extends the scope to a more practical setting by further encompassing no-target and multi-target expressions. Existing REC methods face challenges in handling the complex cases encountered in GREC, primarily due to their fixed output and limitations in multi-modal representations. To address these issues, we propose a Hierarchical Alignment-enhanced Adaptive Grounding Network (HieA2G) for GREC, which can flexibly deal with various types of referring expressions. First, a Hierarchical Multi-modal Semantic Alignment (HMSA) module is proposed to incorporate three levels of alignments, including word-object, phrase-object, and text-image alignment. It enables hierarchical cross-modal interactions across multiple levels to achieve comprehensive and robust multi-modal understanding, greatly enhancing grounding ability for complex cases. Then, to address the varying number of target objects in GREC, we introduce an Adaptive Grounding Counter (AGC) to dynamically determine the number of output targets. Additionally, an auxiliary contrastive loss is employed in AGC to enhance object-counting ability by pulling in multi-modal features with the same counting and pushing away those with different counting. Extensive experimental results show that HieA2G achieves new state-of-the-art performance on the challenging GREC task and also the other 4 tasks, including REC, Phrase Grounding, Referring Expression Segmentation (RES), and Generalized Referring Expression Segmentation (GRES), demonstrating the remarkable superiority and generalizability of the proposed HieA2G.
\end{abstract}

\section{Introduction}
Generalized Referring Expression Comprehension (GREC) \cite{he2023grec,liu2023gres,wu20243d} aims to detect an arbitrary number of target objects based on a given free-form text expression. In contrast to the classic Referring Expression Comprehension (REC) \cite{mao2016generation, yu2016modeling} that only supports the single-target text expressions, GREC narrows the gap with real-world scenarios by further encompassing no-target expressions that do not match any object in the image, and multi-target expressions that refer to multiple target objects. This task has great potential value for various practical applications such as visual-language navigation, embodied AI, and human-robot interaction.

\begin{figure}[t]
    \centering
    \includegraphics[width=0.96\linewidth]{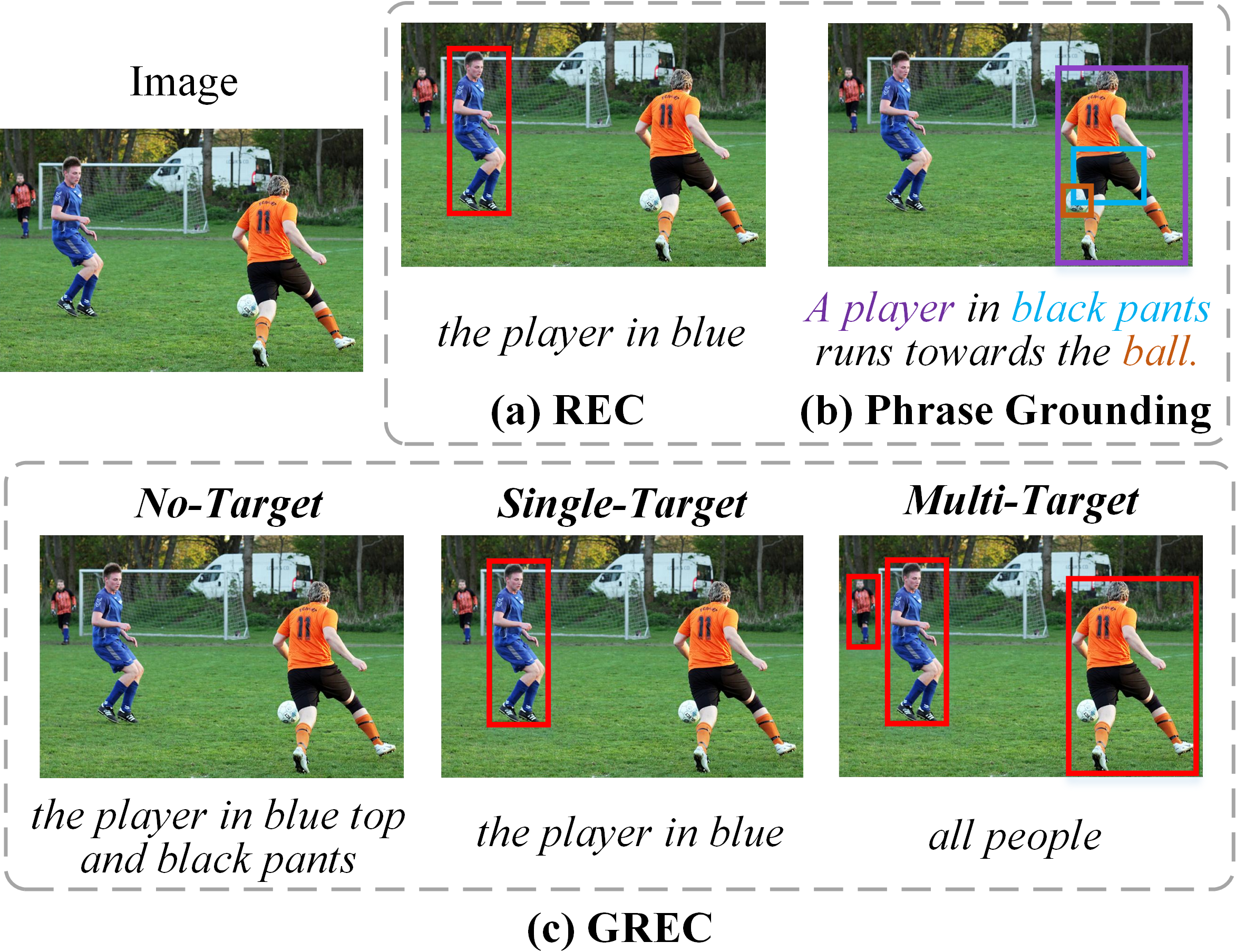}
    \vspace{-3mm}
    \caption{Different visual grounding tasks. 
    (a) Classic REC: text expressions can only specify a single object; (b) Phrase grounding detects all objects mentioned in expressions; (c) GREC~\cite{he2023grec,liu2023gres} supports the text expressions indicating an arbitrary number of target objects from 0 to multiple, which is a more challenging task.}
    \label{fig:fig_example}
    \vspace{-3mm}
\end{figure}

Although recent methods \cite{zhu2022seqtr, deng2023transvg} have achieved remarkable performance in REC, they are constrained to predict only one target that is most related to the text expression, leading to the inability to deal with no-target and multi-target expressions. As shown in Figure \ref{fig:fig_example},  the text expression ``\textit{the player in blue top and black pants}'' does not match any object within the image. In this case, the traditional REC models \cite{kamath2021mdetr, yan2023universal, li2021referring, luo2020multi} still produce a false-negative bounding box. When given the multi-target text expression ``\textit{all people}'', existing REC models also fail to locate all matched targets in the image, arising from the fact that they are enforced to locate only a single target most related to the text expression. Despite phrase grounding \cite{plummer2015flickr30k, yu2020cross} can locate multiple objects, it tends to locate all objects based on the key noun phrases in the text expression without comprehending the entire text semantics. 
GREC is a more challenging task that requires a comprehensive understanding of the intricate semantics of text expressions and visual contents to handle any quantity of target objects ranging from zero to multiple. Therefore, it is necessary to advance a robust GREC model to adapt to this kind of complex generalized scenario.

The first challenge of GREC lies in how to effectively align the diverse text expressions with the corresponding images for comprehensive multi-modal understanding. Existing methods \cite{kamath2021mdetr, radford2021learning, liu2024primitivenet, xu2022groupvit, liu2022open} have made significant efforts to alleviate the cross-modal semantic gap. Nevertheless, they tend to rely solely on single-level alignment, either word-object or text-image alignment, leading to insufficient vision-language interaction and further inhibiting the effective learning of exhaustive multi-modal information. For one image, diverse and flexible text expressions can specify different numbers of target objects from various perspectives as shown in Figure \ref{fig:fig_example}, highlighting the significance of multi-level cross-modal interactions. For example, for a no-target case, the model is often required to capture the fine-grained attribute details of local objects and have a comprehensive understanding of the global contextual information, further rejecting providing any object response. Therefore, it is far from satisfactory to handle the complex cases in GREC leveraging the single-level alignment between the flexible text expressions and images.

The second challenge of GREC is how to output different numbers of target objects dynamically for each specific image-text pair. 
Given a complex text expression specifying multiple targets, a potential approach is to split the text expression into multiple text expressions and query the model multiple rounds to obtain the target objects one by one. However, text expressions with implicit multi-target information are difficult to decompose, and such an approach can not solve the inherent requirement in GREC, which desires an efficient model to give all targets in a single forward process. More importantly, multi-target expressions such as ``\textit{three players}'' and ``\textit{all people}'' necessitate a model to possess an explicit or implicit object-counting ability. Although a threshold-based strategy \cite{he2023grec} has demonstrated its advantages in selecting the target objects from multiple candidate object proposals, it is often challenging to decide an appropriate threshold for all samples. Moreover, using a unified threshold struggles to adapt to the characteristics of different samples, resulting in inaccurate prediction results for some samples. Therefore, it is crucial to design a more advanced strategy for selecting target objects.

To address the above challenges, we propose \textbf{HieA2G}, a \textbf{Hie}rarchical \textbf{A}lignment-enhanced~\textbf{A}daptive~\textbf{G}rounding Network, for GREC to deal with various types of referring expressions flexibly. Specifically, we design a Hierarchical Multi-modal Semantic Alignment (HMSA) module to achieve comprehensive and robust multi-modal understanding by coupling three levels of alignments including word-object, phrase-object, and text-image alignment. Due to the absence of fine-grained region-level annotations corresponding to the entity words, we propose a text mask recovery auxiliary task to reconstruct the masked text semantics with the visual object features to promote word-object alignment. In this way, the visual features are facilitated to fuse the semantic information of the masked entity fully. Compared with the entity word, the attribute-related information is essential to distinguish the object of the same category. By matching the descriptive phrase with the visual object, the encoder can derive distinctive visual features and understand a larger range of semantic units. After that, the high-level text-image alignment matches the overall semantics between the text and image, enabling a more comprehensive perception of global information. 
On the one hand, the HMSA module can help provide holistic and robust multi-modal understanding to facilitate more accurate object localization. On the other hand, it endows the model with the ability to exploit information at various levels of detail, which allows it to accomplish other various tasks like REC and phrase grounding. Furthermore, to address the varying number of target objects in GREC, we design an Adaptive Grounding Counter (AGC) to dynamically determine the number of output target objects for each specific image-text pair. Additionally, an auxiliary contrastive loss is employed in AGC to enhance the object-counting ability by pulling together the multi-modal features with the same counting and pushing away those with different counting.

Our contributions are summarized as follows: 
(1) We propose a Hierarchical Alignment-enhanced Adaptive Grounding Network (HieA2G) for GREC to support text expressions indicating an arbitrary number of target objects.  (2) We design a Hierarchical Multi-modal Semantic Alignment module to enable hierarchical vision-language interactions across multiple levels for comprehensive and robust multi-modal semantic understanding.  (3) We propose an Adaptive Grounding Counter to dynamically determine the number of output targets for each specific image-text pair, which can help deal with the multi/single/no-target text expressions flexibly.  (4) Extensive experimental results show that HieA2G achieves new SOTA results on the challenging GREC task. It also exhibits superior performance across the other four visual grounding tasks including REC, Phrase Grounding, Referring Expression Segmentation (RES), and Generalized Referring Expression Segmentation (GRES).

\begin{figure*}[t]
    \centering
  \includegraphics[width=0.92\linewidth]{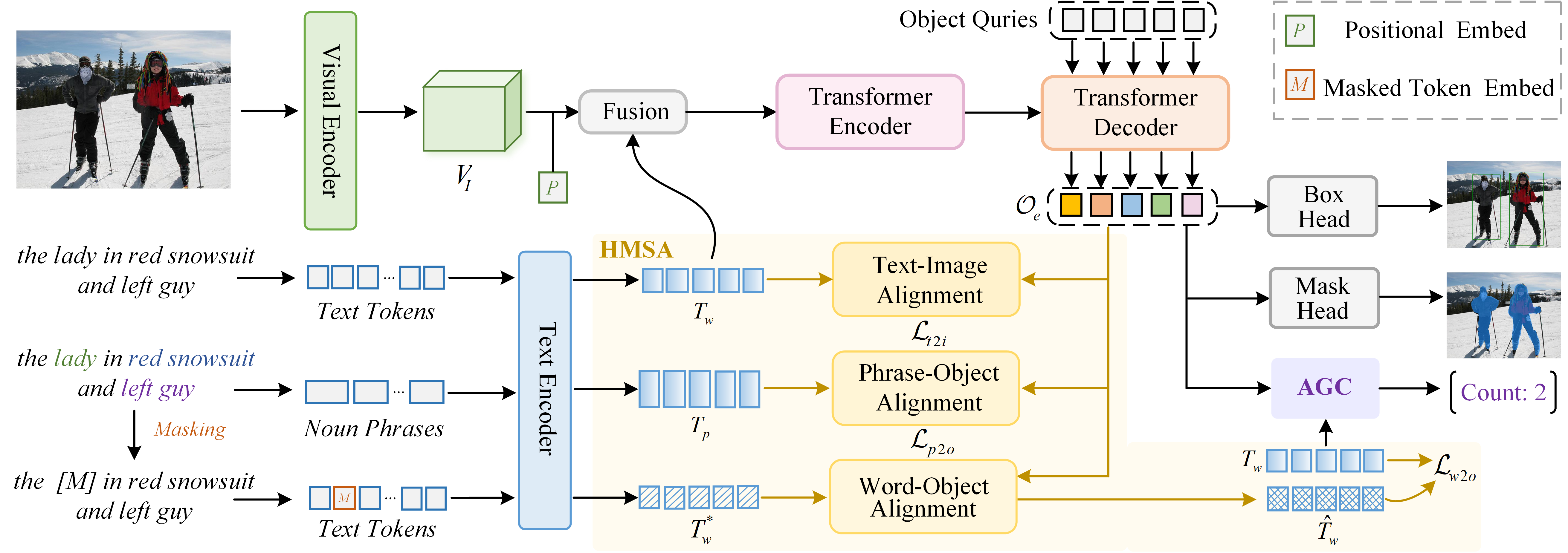}
  \vspace{-3mm}
    \caption{The framework of our proposed HieA2G. First, the visual encoder and the text encoder extract the visual feature $V_I$ and text feature $T_w$. Then, a Transformer encoder is employed to perform multi-modal feature interaction further. The learnable object queries and the output of the Transformer encoder are fed to the Transformer decoder, whose output is object embeddings $\mathcal{O}_e$ corresponding to the object queries. Next, based on  $\mathcal{O}_e$, the Hierarchical Multi-modal Semantic Alignment (HMSA) module is employed to facilitate multi-level cross-modal interaction via word-object, phrase-object, and text-image alignment. Moreover, an Adaptive Grounding Counter (AGC) is utilized to decide the output number of target objects dynamically.}
    \label{fig:model}
    \vspace{-3mm}
\end{figure*}

\section{Related Work}
\textbf{Referring Expression Comprehension.}
REC aims to detect one specific object from an image based on a referring expression. Existing methods can be classified into two groups: two-stage \cite{hu2017modeling, zhuang2018parallel, yang2019dynamic, liu2019improving, li2021bottom}  and one-stage \cite{liao2020real, zhou2021real, yang2022improving, deng2023transvg, ye2021one} methods. The recent advancements in large language models \cite{WUPAMI, chen2023shikra, youferret} have also
brought new opportunities to vision-language tasks requiring localization like REC. They have achieved promising results by collecting large-scale datasets, pretraining, and fine-tuning the large language models. It’s worth noting that our work focuses on GREC, which detects an arbitrary number of target objects. Therefore, the top-1 selection method for REC cannot be applied directly in this setting. Although GREC \cite{he2023grec} has modified some REC methods \cite{luo2020multi, kamath2021mdetr, ding2022vlt, yan2023universal} to output different numbers of bounding boxes, they still struggle to deal with such complex and flexible referring expressions, leading to unsatisfactory performance.

\noindent{\textbf{Referring Expression Segmentation.}}
RES aims to segment one object based on text expression~\cite{liu2022instance}. Driven by the success of Transformer, recent works \cite{ding2021vision, MeViS, MOSE, ding2022vlt, RefMask3D, SegPoint, WUCVPR, TransformerSurvey, wu2024towards, OpenVocabularySurvey, liu2023multi, liu2024referring, li2021referring, kim2022restr, yang2022lavt, wang2022cris} have extensively use it to extract visual and language features. ReLA \cite{liu2023gres} introduces the Generalized Referring Expression Segmentation (GRES) benchmark, which further include multi-target and no-target samples. They only study region-language and region-image relationships. In contrast, we propose a hierarchical multi-modal alignment to enhance the comprehensive understanding of visual-linguistic context. Moreover, rather than a binary classification to judge only the existence of objects, we design an adaptive object-counting strategy to facilitate robust object perception.

\section{Methodology}
\subsection{Architecture Overview}
Figure \ref{fig:model} shows the overall architecture of our proposed HieA2G. First, the text expression $T$ is fed into the text encoder to obtain the word features $T_w = \{w_k | k \in \{1, 2, ..., K\}\}$, where $K$ is the number of words. For the phrase feature, we first extract noun phrases from the text expression, which are then represented as $T_p = \{p_i | i \in \{1, 2, ..., M\}\}$ by average pooling the word features in each phrase, where $M$ is the number of phrases. For the image input $I$, we adopt a visual encoder to obtain the visual feature $V_I$ and flatten it into a 2D feature combined with positional embeddings. The image feature and text feature are projected into the same space, and then are fused by concatenation to fed to the Transformer encoder for multi-modal deep fusion. Next, the output of the Transformer encoder and the learnable object queries are fed into the Transformer decoder. Subsequently, we obtain the text-aware object embeddings $\mathcal{O}_e = \{o_j | j \in \{1, 2, ..., N\}\}$ corresponding to the $N$ object queries. Based on the object embeddings $\mathcal{O}_e$, the Hierarchical Multi-modal Semantic Alignment (HMSA) module is performed to hierarchically incorporate the multi-modal information across multiple levels for a comprehensive multi-modal understanding. Furthermore, we propose an Adaptive Grounding Counter (AGC) to determine the output number of target objects dynamically and then select the desired outputs from candidate object proposals.

\subsection{Hierarchical Multi-modal Semantic Alignment} \label{3.2}
To fully model the relationships between the various types of text expression and the images, we propose a Hierarchical Multi-modal Semantic Alignment (HMSA) to facilitate multi-level cross-modal interactions through word-object, phrase-object, and text-image alignment. HMSA can help exploit information at various levels of detail and promote comprehensive and robust text-aware object embeddings for better box regression and mask segmentation.

\subsubsection{Word-Object Alignment.} 
To enable a more directly fine-grained word-object alignment, we introduce a masked text recovery task by enforcing the model to recover the missed key information in the text based on the matched object features. The object embeddings are then facilitated to fuse the semantic information of the masked entity fully. Specifically, we first extract the entity noun in the text and randomly mask it with a [MASK] token. The masked text is encoded by the text encoder as $T_w^{*}$. Then, combining the object embeddings $\mathcal{O}_e$ and the masked text feature $T_w^{*}$, a Transformer layer is utilized to reconstruct the text semantic feature $\hat{T}_w$. 
We design a masked text recovery loss $\mathcal{L}_{w2o}$ by measuring the semantic similarity between the original complete text feature $T_w$ and the reconstructed text feature $\hat{T}_w$ as:
\begin{equation}
    \mathcal{L}_{w2o} = \alpha(1-\text{cos}(T_w, \hat{T}_w)),
\end{equation}
where $\alpha$ is set to 0 when given a no-target sample, otherwise set to 1.  Due to the weak relevance and even total irrelevance between the textual and the visual features for no-target samples, it is meaningless and impossible to reconstruct the missing information for this kind of image-text pair, even interfering with model optimization.

\subsubsection{Phrase-Object Alignment.}
With the explicit phrase-object annotations in the Flickr30K Entities \cite{plummer2015flickr30k} dataset, it is promising to encourage the model to match each phrase with the corresponding object query. 
By matching the descriptive phrase with the visual object, we can derive distinctive object features and understand a larger range of semantic units. Specifically, we first project the phrase features $T_p \in \mathbb{R}^{M \times C_p}$ and the object embeddings $\mathcal{O}_e \in \mathbb{R}^{N \times C_v}$ into the same sub-space by linear layers:
\begin{equation}
   \widehat{T}_p = W_1 T_p, \quad \widehat{\mathcal{O}}_e = W_2 \mathcal{O}_e,
\end{equation}
where $W_1$ and $W_2$ are learnable parameters, $\widehat{T}_p  \in \mathbb{R}^{M \times C}$ and $\widehat{\mathcal{O}}_e \in \mathbb{R}^{N \times C}$ are the projected phrase features and projected object embeddings, $M$ and $N$ is the number of phrases and object queries, respectively. The matching relation map $S \in \mathbb{R}^{M \times N}$ between all noun phrases and the object queries is calculated as follows:
\begin{equation}
   S = {\rm{Sigmoid}}(\widehat{T}_p\cdot \widehat{\mathcal{O}}_e^\top).
\end{equation}

For all object queries, we adopt a bipartite matching \cite{cheng2022masked} to find the matched ground-truth bounding box. Then, we can obtain a ground-truth binary map $Y \in \mathbb{R}^{M \times N}$ between phrases and object queries, indicating their matching relationships. With the predicted matching relation map $S \in \mathbb{R}^{M \times N}$, we design a phrase-object contrastive loss $\mathcal{L}_{p2o}$, implemented with the binary cross-entropy loss:
\begin{equation}
   \mathcal{L}_{p2o} = -\sum_{i=1}^{M} \sum_{j=1}^{N} Y_{i,j}{\rm{log}}S_{i,j} + (1-Y_{i,j}){\rm{log}}(1-S_{i,j}).
\end{equation}

In this way, the model is encouraged to generate higher scores for positive phrase-object alignments and lower scores for negative ones. Therefore, the object embeddings can be endowed with stronger discriminative ability by capturing fine-grained attributes within the phrase semantics.

\subsubsection{Text-Image Alignment.} 
The image feature should have high feature similarity with the matched text expression and low feature similarity with the unmatched text expression. To fully model the global relationships between the image $I$ and text expression $T$, we define a global match score $S^T(I, T)$ for each image-text pair to calculate their similarity via word-object pairs as follows:
\begin{align}
   S^T(I, T) &= \frac{1}{N}\sum_{j=1}^{N} \sum_{k=1}^{K}a_{j,k} \langle \hat{o}_j, \hat{w}_k \rangle, \\
   a_{j,k} &= \frac{\text{exp}\langle \hat{o}_j, \hat{w}_k \rangle}{ \sum_{l=1}^{K}{\text{exp}\langle \hat{o}_j, \hat{w}_l \rangle}},
\end{align}
where $\langle . \ , . \rangle$ represents the dot product operation of two embeddings, $\hat{o}
_j$ and $\hat{w}_k$ denotes the projected embedding of $j$-th object query and $k$-th word, and $S^T(I, T)$ is computed by normalizing along the text dimension. Similarly, $S^I(I, T)$ can be obtained by normalizing along the image dimension.

The global match score $S^T(I, T)$ measures the degree of semantic alignment between an image and its corresponding text expression. In this case, maximizing the match score of matched image-text pairs helps ensure their strong correspondence. For an image-text pair in a batch, the objective function is defined as follows:
\begin{align}
  \mathcal{L}^{TT}_{t2i}(I) = -\log \frac{\exp(S^T(I, T))}{\sum_{T' \in \mathcal{B}_T} \exp(S^T(I, T'))}, \\
    \mathcal{L}^{TI}_{t2i}(T) = -\log \frac{\exp(S^T(I, T))}{\sum_{I' \in \mathcal{B}_I} \exp(S^T(I', T))}, 
\end{align}
where $\mathcal{B}_T$, $\mathcal{B}_I$ represents a collection of the text expressions and images in a batch, $\mathcal{L}^{TT}_{t2i}(I)$ and $\mathcal{L}^{TI}_{t2i}(T)$ are normalized along text and image dimension, respectively. Similarly, $\mathcal{L}^{IT}_{t2i}(I)$ and  $\mathcal{L}^{II}_{t2i}(T)$ can be obtained using $S^I(I, T)$. 

The final text-image alignment loss for each image-text pair is computed as follows:
\begin{equation}
    \mathcal{L}_{t2i} = \mathcal{L}^{TT}_{t2i}(I) + \mathcal{L}^{IT}_{t2i}(I))+\mathcal{L}^{TI}_{t2i}(T) + \mathcal{L}^{II}_{t2i}(T),
\end{equation}

In this way, the thorough text-image alignment boosts a more comprehensive understanding of global multi-modal information.

The final loss of HMSA is computed as $\mathcal{L}_{align} =  \mathcal{L}_{w2o} + \mathcal{L}_{p2o} + \mathcal{L}_{t2i}$.
By incorporating these three-level alignments, the object embeddings corresponding to the object queries can be gradually refined for better box regression and mask segmentation, further obtaining $N$ potential object proposals via box head and mask head.

\subsection{Adaptive Grounding Counter} \label{3.3}

To adapt to the generalized setting such as no-target and multi-target samples, we design an Adaptive Grounding Counter (AGC) to decide the output number of target objects dynamically for each specific image-text pair. With AGC, the desired target objects can be selected effectively from the $N$ candidate object proposals corresponding to the object queries. Specifically, we formulate it as a classification task to predict an output number. For different images, the same text expression can specify different target objects. For example, ``\textit{all kids}'' can denote an arbitrary number of objects in different images. Therefore, it is necessary to incorporate both the sentence feature of the text expression and object embeddings to predict the label. To better train AGC, we conducted a statistical analysis of all samples in the gRefCOCO dataset \cite{liu2023gres, he2023grec}. The distribution of objects follows a long-tailed pattern, with most samples falling within the range of 0 to 3, and only a small proportion exceeding the number of 3.

\begin{figure}
    \centering
    \includegraphics[width=1\linewidth]{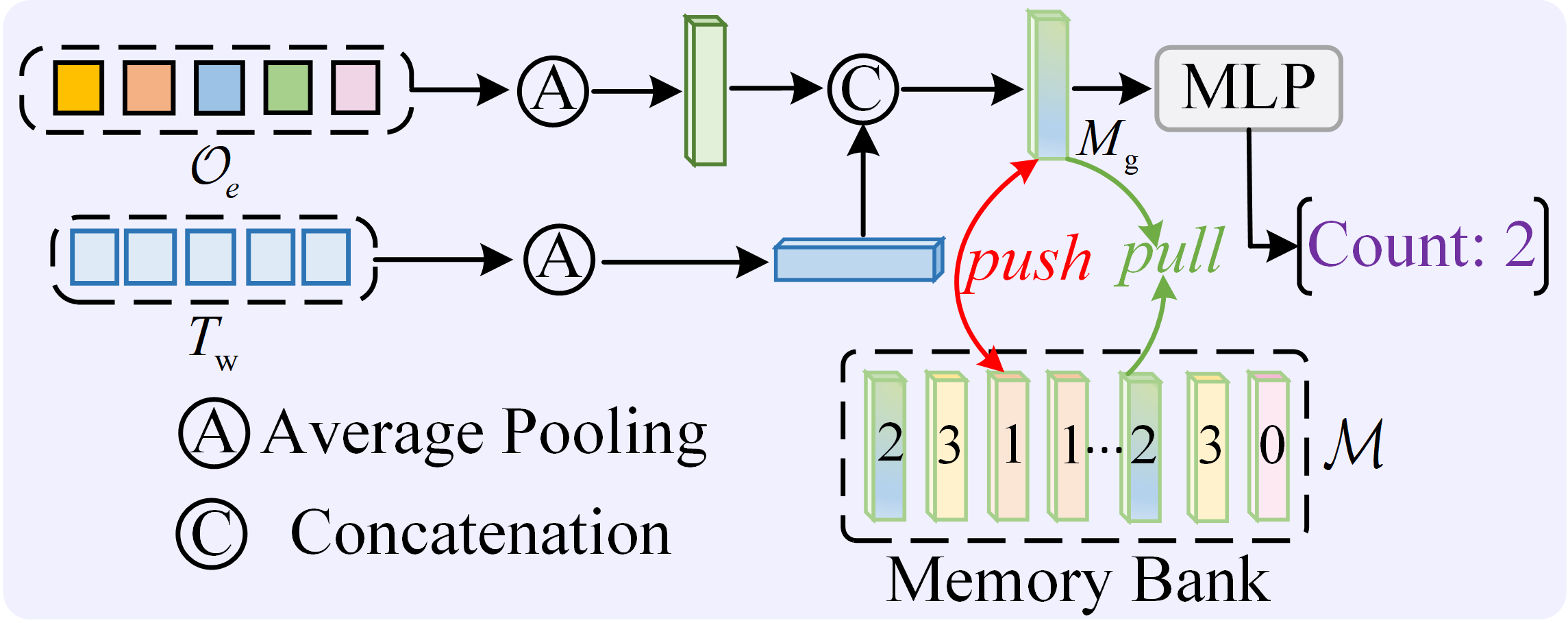}
    \caption{The detail of the Adaptive Grounding Counter.}
    \label{fig:deatail}
\end{figure}

Therefore, it is defined as a classification task into five classes. With word features $T_w$ and object embeddings $\mathcal{O}_e$ at hand, we adopt the average pooling to obtain the global text feature and visual feature. Then, the two features are concatenated to obtain a global multi-modal feature $M_g$, which is used to predict the object counting label $y_c$ as:
\begin{align}
    M_g &= [\text{AP}(T_w); \text{AP}(\mathcal{O}_e)], \\
    y_c &= {\rm{MLP}}(M_g),
\end{align}
where [;] denotes concatenation operation, AP denotes average pooling, MLP denotes a two-layer perceptron, and $y_c\in\{0, 1, 2, 3, 3+\}$. Note that, only when the counting is larger than 3, the threshold-based strategy is adopted, that is object proposals with class scores above the threshold are selected. Otherwise, the sorted target objects with high scores are selected according to the counting.

To further promote the object counting ability, we incorporate contrastive learning in AGC by pulling together the multi-modal features $M_g$ with the same counting and pushing away those with the different counting in Figure \ref{fig:deatail}. The number of negative samples is related to batch size. However, the size of the batch size is limited by GPU memory. Therefore, to facilitate contrastive learning, we introduce a memory bank $\mathcal{M}$ \cite{he2020momentum} to maintain a larger number of negative samples. Inspired by \cite{khosla2020supervised}, a supervised contrastive loss $\mathcal{L}_{con}$ is introduced as follows: 
\begin{equation}
 \small \mathcal{L}_{con} = -\frac{1}{\lvert P(i) \rvert} \sum_{p\in P(i)} \text{log} \frac{\text{exp}(M_g^i \cdot M_g^p / \tau)}{\sum_{a\in A(i)} \text{exp}(M_g^i\cdot M_g^a / \tau)},
\end{equation}
where $i$ denotes anchor index, $P(i)=\{p \in A(i), y_c^p=y_c^i\}$ is a collection of indices for the positive samples in $\mathcal{M}$, $|P(i)|$ denotes the cardinality of the collection, $A(i)$ denotes a collection of indices for all positive and negative samples in $\mathcal{M}$, $M_g$ denotes the global multi-modal feature, and $\tau$ is a temperature hyperparameter.

The final loss of AGC is computed as $\mathcal{L}_{agc} =  \mathcal{L}_{cls} + \mathcal{L}_{con}$, where $\mathcal{L}_{cls}$ denotes the object counting classification loss, implemented by a cross-entropy loss.

\subsection{Training Objective}
To further supervise task-specific training, a series of losses for the box head and mask head are introduced as follows:
\begin{align}
   \mathcal{L}_{det} &= \lambda_{bbox} \mathcal{L}_{bbox} + \lambda_{giou} \mathcal{L}_{giou} + \lambda_{class} \mathcal{L}_{class},\\
   \mathcal{L}_{seg} &= \lambda_{mask}  \mathcal{L}_{mask} + \lambda_{dice} \mathcal{L}_{dice},
\end{align}
where $\lambda_{*}$ are the hyperparameters, $\mathcal{L}_{class}$ is cross-entropy loss for box classification, $\mathcal{L}_{bbox}$ and $\mathcal{L}_{giou}$ are L1 loss \cite{ren2015faster} and GIoU loss \cite{rezatofighi2019generalized} for box regression. The focal loss $\mathcal{L}_{mask}$ \cite{lin2017focal} and dice loss $\mathcal{L}_{dice}$ \cite{milletari2016v} are introduced to supervise mask segmentation \cite{CCL}.

We first pretrain HieA2G on a combined dataset formed by the training data of RefCOCO/+/g, Flickr30K Entities, and gRefCOCO datasets using the joint loss $\mathcal{L}_{pretrain} = \mathcal{L}_{align} + \mathcal{L}_{det}$. The goal of pretraining is to incorporate comprehensive multi-modal information into object queries. Then based on the pretrained weights, HieA2G is finetuned on various downstream tasks with the task-specific loss such as $\mathcal{L}_{det}$, $\mathcal{L}_{mask}$ and an additional loss $\mathcal{L}_{agc}$ introduced specifically for GREC and GRES.

\begin{table}[t]
\centering
\footnotesize
\setlength{\tabcolsep}{1.7mm}
{\renewcommand{\arraystretch}{1}
\begin{tabular}{l|cc|cc|cc}
\hline
\multirow{2}{*}{Methods}       & \multicolumn{2}{c|}{val}  & \multicolumn{2}{c|}{testA} & \multicolumn{2}{c}{testB} \\
            & Pr & N-acc. & Pr  & N-acc. & Pr & N-acc. \\ \hline
MCN$^{\dag}$              & 28.0              & 30.6  & 32.3               & 32.0  & 26.8              & 30.3  \\
VLT$^{\dag}$                & 36.6              & 35.2  & 40.2               & 34.1  & 30.2              & 32.5  \\
MDETR$^{\dag}$          & 42.7              & 36.3  & 50.0              & 34.5  & 36.5              & 31.0  \\
UNITEXT$^{\dag}$          & 58.2              & 50.6  & 46.4               & 49.3  & 42.9              & 48.2   \\ 
Ferret$^{*}$        & 54.8             & 48.9                 & 49.5   & 45.2              & 43.5  & 43.8   \\ 

\hline
\rowcolor{gray!15} \textbf{HieA2G}$_\textrm{R101}$         & \textbf{67.8}                  & \textbf{60.3}       &  \textbf{66.0}                   & \textbf{60.1}       & \textbf{56.5}                   &   \textbf{56.0}  
\\ 
\hline
\end{tabular}}
\caption{Results on gRefCOCO dataset \cite{liu2023gres} in terms of Pr@($\mathrm{F_1}$=1, IoU$\geq$0.5) and N-acc. for GREC task. ${\dag}$ denotes these methods have been modified to generate multiple boxes following \cite{he2023grec}.  
${\ast}$ denotes the model adapted for the GREC task.}
\label{GREC}
\end{table}

\section{Experiments}
\subsection{Experimental Setup}
\textbf{Datasets.} The proposed HieA2G is evaluated mainly on the gRefCOCO dataset \cite{he2023grec, liu2023gres} for GREC and GRES. We also conducted experiments on a phrase grounding dataset called Flickr30K Entities \cite{plummer2015flickr30k}, and three widely-used REC and RES benchmarks including RefCOCO \cite{yu2016modeling}, RefCOCO+ \cite{yu2016modeling}, and RefCOCOg \cite{mao2016generation}. 

\noindent \textbf{Implementation Details.}
We adopt ResNet101 \cite{he2016deep} and Swin-B \cite{liu2021swin} as our visual encoder, and RoBERTa-base \cite{liu2019roberta} as our text encoder. 

\begin{table}[t]
\centering
\footnotesize
\setlength{\tabcolsep}{0.9mm}
{\renewcommand{\arraystretch}{1}
\begin{tabular}{l|ccc|ccc|cc}
\hline
\multirow{2}{*}{Methods} &  \multicolumn{3}{c|}{RefCOCO} & \multicolumn{3}{c|}{RefCOCO+} & \multicolumn{2}{c}{RefCOCOg} \\
                                                 & val      & testA   & testB   & val      & testA    & testB   & val-u        & test-u        \\ \hline
MAttNet                                   & 76.7    & 81.1   &  70.0   &  65.3   &  71.6    & 56.0   &  66.6       & 67.3        \\    
RefTR                                 & 85.7    & 88.7   & 81.2   & 77.6    & 82.3    & 69.0   & 79.3        & 80.0        \\
MDETR                             & 86.8    & 89.9   & 81.4   & 79.5    & 84.1    & 70.6 & 81.6        & 80.9         \\
SeqTR      
                 & 83.7   & 86.5   & 81.2   & 71.5    & 76.3    & 64.9   & 74.9       & 74.2         \\ 
TransVG++                         & 86.3   & 88.4  & 81.0   & 75.4   & 80.5    & 66.3   & 76.2           & 76.3    \\
LISA-7B                                  & 85.4  & 88.8    &  82.6    & 74.2   & 79.5       & 68.4    & 79.3   & 80.4        \\ 
GSVA-7B                                  & 86.3  & 89.2    & 83.8    & 72.8   & 78.8       & 68.0    & 81.6   & 81.8        \\ \hline
\rowcolor{gray!15} \textbf{HieA2G}$_\textrm{R101}$                  & \textbf{87.8}            & \textbf{90.3}                & \textbf{84.0}      & \textbf{80.7}                   & \textbf{85.6}      & \textbf{72.9}                 & \textbf{83.7} & \textbf{83.8}       \\ 
\hline
\end{tabular}}
\caption{Results comparison on RefCOCO/+/g for REC task.}
\label{REC}
\end{table}

\begin{table}[t]
\centering
\fontsize{9pt}{10pt}\selectfont
\setlength{\tabcolsep}{1.66mm}
{\renewcommand{\arraystretch}{1}
\begin{tabular}{l|ccc|ccc}
\hline
\multirow{2}{*}{Methods} & \multicolumn{3}{c|}{val} & \multicolumn{3}{c}{test} \\
                        & R@1    & R@5    & R@10  & R@1    & R@5    & R@10   \\ \hline
VisualBert            & 68.1   & 84.0   & 86.2  & -        & -       & -       \\
VisualBert              & 70.4   & 84.5   & 86.3  & 71.3   & 85.0   & 86.5   \\
MDETR                 & 82.5   & 92.9   & 94.9  & 83.4   & 93.5   & 95.3   \\ 
Shrika-7B                  &  75.8  & -   & -  & 76.5   & -  & -   \\
Ferret-7B               &  80.4  & -   & -  & 82.2   & -  & -   \\ \hline
\rowcolor{gray!15} \textbf{HieA2G}$_\textrm{R101}$                
                        & \textbf{82.9}       & \textbf{93.2}       & \textbf{95.1}     &\textbf{83.7}        & \textbf{93.8}       & \textbf{95.6}      \\ \hline
\end{tabular}}
\caption{Results comparison on Flickr30K Entities dataset in Recall@k (ANY-BOX protocol) for Phrase Grounding task.}
\label{phrase}
\end{table}

\begin{table}[t]
\centering
\footnotesize
\setlength{\tabcolsep}{0.8mm}
{\renewcommand{\arraystretch}{1}
\begin{tabular}{p{1.7cm}|ccc|ccc|cc}
\hline
\multirow{2}{*}{Methods} &  \multicolumn{3}{c|}{RefCOCO} & \multicolumn{3}{c|}{RefCOCO+} & \multicolumn{2}{c}{RefCOCOg} \\
                       &                            val      & testA   & testB   & val      & testA    & testB   & val-u        & test-u        \\ \hline

MAttNet                                                & 56.5 & 62.4 & 51.7 & 46.7 & 52.4 & 40.1 & 47.6 & 48.6 \\

MCN                                                    & 62.4 & 64.2 & 59.7 & 50.6 & 55.0 & 44.7 & 49.2 & 49.4 \\
VLT                                                   & 65.7 & 68.3 & 62.7 & 55.5 & 59.2 & 49.4 & 52.9 & 56.7 \\
\rowcolor{gray!15} \textbf{HieA2G}$_\textrm{R101}$                   & \textbf{73.3}          & \textbf{75.9}               &  \textbf{69.0}     & \textbf{64.8}                  & \textbf{69.7}      &  \textbf{56.1}                 & \textbf{62.9}  & \textbf{63.5}      \\ 
\hline
LAVT                                                      & 72.7 & 75.8 & 68.8 & 62.1 & 68.4 & 55.1 & 61.2 & 62.1 \\ 
ReLA                                                      & 73.8 & 76.5  & 70.2  & 66.0 & 71.0 & 57.7 & 65.0 & 66.0  \\ 
LISA-7B                                                   & 74.9  & \textbf{79.1}  &  72.3  & 65.1 & 70.8 & 58.1 & 67.9 & 70.6 \\ 
GSVA-7B                                                    & \textbf{77.2}  & 78.9  &  \textbf{73.5}  & 65.9 & 69.6 & \textbf{59.8} & \textbf{72.7} & \textbf{73.3} \\

\rowcolor{gray!15} \textbf{HieA2G}$_\textrm{SwinB}$                   & 75.1          & 77.6              & 71.1    & \textbf{66.5}                  & \textbf{71.4}      &  58.9                & 65.3  & 66.6     \\
\hline
\end{tabular}}
\caption{Results comparison on RefCOCO/+/g for RES task.}
\label{RES}
\end{table}

\begin{figure*}[t]
    \centering
\includegraphics[width=0.94\linewidth]{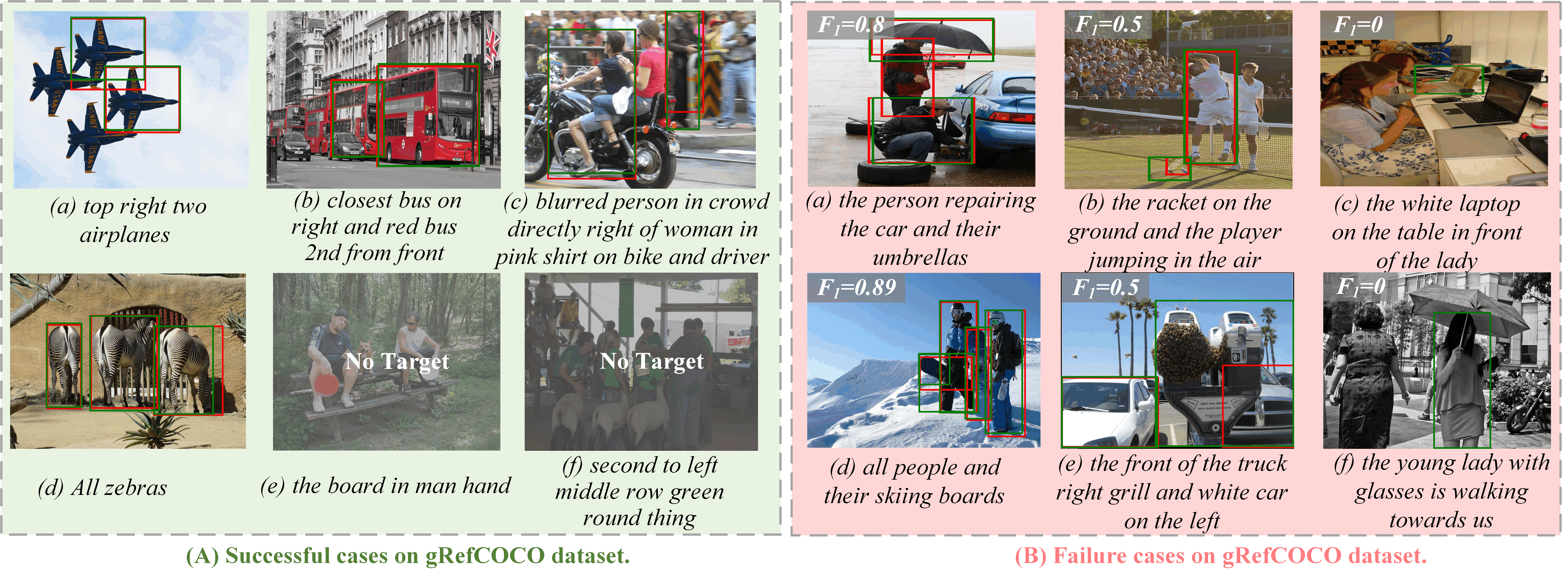}
    \caption{Visualization for the success cases and failure cases of HieA2G on gRefCOCO dataset. The ground truth is denoted by red bounding boxes, whereas green bounding boxes denote the predictions. The $\mathrm{F_1}$ score of all success cases in (A) is 1.0.}
    \label{fig:fig_case}
\end{figure*}

\begin{table*}[]
\footnotesize
\begin{tabular}{l|c|cccc|cccc|cccc}
\hline
\multirow{2}{*}{Methods}                                             & \multirow{2}{*}{Backbone} & \multicolumn{4}{c|}{val}                                             & \multicolumn{4}{c|}{testA}                                           & \multicolumn{4}{c}{testB}                       \\
                                                                     &                           & cIoU          & gIoU                               & N-acc. & T-acc. & cIoU          & gIoU                               & N-acc. & T-acc. & cIoU          & gIoU          & N-acc. & T-acc. \\ \hline
MAttNet                                                              & ResNet101                     & 47.5          & \multicolumn{1}{c}{48.2}          & 41.2   & 96.1   & 58.7          & \multicolumn{1}{c}{59.3}          & 44.0   & 97.6   & 45.3          & 46.1          & 41.3   & 95.3   \\
VLT                                                                  & DarkNet53                      & 52.5          & \multicolumn{1}{c}{52.0}          & 47.2   & 95.7   & 62.2          & \multicolumn{1}{c}{63.2}          & 48.7   & 95.9   & 50.5          & 50.9          & 47.8   & 94.7   \\
VLT+ReLA                                                             & DarkNet53                      & 58.7          & \multicolumn{1}{c}{59.4}          & -      & -      & 66.6          & \multicolumn{1}{c}{65.4}          & -      & -      & 56.2          & 57.4          & -      & -      \\
CRIS                                                                 & ResNet101                   & 55.3          & \multicolumn{1}{c}{56.3}          & -      & -      & 63.8          & \multicolumn{1}{c}{63.4}          & -      & -      & 51.0          & 51.8          & -      & -      \\
\rowcolor{gray!15} \textbf{HieA2G}    & ResNet101                     & \textbf{62.5} & \multicolumn{1}{c}{\textbf{67.1}} & \textbf{60.9}   & \textbf{97.4}   & \textbf{67.6} & \multicolumn{1}{c}{\textbf{70.5}} & \textbf{60.2}   & \textbf{97.7}   & \textbf{58.8} & \textbf{61.5} & \textbf{56.5}   & \textbf{96.4}   \\ \hline
LAVT                                                                 & Swin-B                    & 57.6          & \multicolumn{1}{c}{58.4}          & 49.3   & 96.2   & 65.3          & \multicolumn{1}{c}{65.9}          & 49.3   & 95.1   & 55.0          & 55.8          & 48.5   & 95.3   \\
ReLA                                                                 & Swin-B                    & 62.4          & \multicolumn{1}{c}{63.6}          & 56.4   & 96.3   & 69.3          & \multicolumn{1}{c}{70.0}          & 59.0   & 97.8   & 59.9          & 61.0          & 58.4   & 95.4   \\
LISA-7B                                                              & ViT-H                     & 61.8          & \multicolumn{1}{c}{61.6}          & 54.7   & -      & 68.5          & \multicolumn{1}{c}{66.3}          & 50.0   & -      & 60.6          & 58.8          & 51.9   & -      \\
GSVA-7B                                                              & ViT-H                     & 63.3          & \multicolumn{1}{c}{66.5}          & 62.4   & -      & 69.9          & \multicolumn{1}{c}{71.1}          & \textbf{65.3}   & -      & 60.5          & 62.2          & 60.6   & -      \\
\rowcolor{gray!15} \textbf{HieA2G}  & Swin-B                    & \textbf{64.2} & \multicolumn{1}{c}{\textbf{68.4}} & \textbf{62.8}   & \textbf{98.3}   & \textbf{70.4} & \multicolumn{1}{c}{\textbf{72.0}} & 63.4   & \textbf{98.5}   & \textbf{61.0} & \textbf{62.8} & \textbf{60.8}   & \textbf{97.5}   \\ \hline
\end{tabular}
\caption{Results comparison on gRefCOCO dataset in terms of cIoU, gIoU, N-acc. and T-acc. for GRES task.}
\label{GRES-1}
\end{table*}

\begin{table}[t]
\centering
\footnotesize
\setlength{\tabcolsep}{1.9mm}
{\renewcommand{\arraystretch}{1}
\begin{tabular}{c|ccc|cc|cc}
\hline
\multirow{2}{*}{\#} & \multicolumn{3}{c|}{HMSA}            & \multicolumn{2}{c|}{AGC}                         & \multicolumn{2}{c}{GREC} \\ 

                  & W2O & P2O & \multicolumn{1}{c|}{T2I} & Classifier & \multicolumn{1}{c|}{$\mathcal{L}_{con}$} & Pr        & N-acc.        \\ \hline

\#1                 &     &     &                          & $\checkmark$       &  $\checkmark$                                  & 65.2       & 54.9          \\      \#2                 &      & $\checkmark$     & $\checkmark$                      & $\checkmark$        &$\checkmark$                                    & 67.5          & 58.1             \\                           
\#3                & $\checkmark$    &   &  $\checkmark$                        & $\checkmark$       & $\checkmark$                                    & 67.1         & 57.3             \\
\#4               & $\checkmark$      & $\checkmark$    &                        & $\checkmark$      &$\checkmark$  & 67.0          & 56.4             \\ \hline
\#5                 & $\checkmark$    & $\checkmark$   & $\checkmark$                      &       &                                  &53.9  &  48.0            \\ 

\#6                & $\checkmark$ &$\checkmark$   & $\checkmark$ & $\checkmark$          &                                      & 66.5 &   57.3    \\ 
\rowcolor{gray!15}  \#7                 & $\checkmark$   & $\checkmark$  & $\checkmark$                     & $\checkmark$      & $\checkmark$                                     & \textbf{67.8}         & \textbf{60.3}                        \\ \hline
\end{tabular}}
\caption{Ablation study of different components of HieA2G for GREC. Notably, W2O, P2O and T2I indicate the word-object,  phrase-object, and text-image alignment.}
\label{aba}
\end{table}

\subsection{Performance Comparison}
\noindent \textbf{Results on GREC.}
As shown in Table \ref{GREC}, our HieA2G with the ResNet101 backbone achieves superior performance on both metrics across three splits of the gRefCOCO dataset. It shows an average performance gain of 14.2\% in Pr@($\mathrm{F_1}$=1, IoU $\geq$ 0.5) over Ferret \cite{youferret} using a Multimodal Large Language Model (MLLM), and an average performance gain of 9.4\% in N-acc. over UNITEXT \cite{yan2023universal}. These results indicate that HieA2G has a significant advantage in handling various types of text expressions to flexibly detect target objects ranging from zero to multiple.

\noindent \textbf{Results on REC.}
As illustrated in Table \ref{REC}, HieA2G achieves consistent performance gains across all splits of the three datasets compared to existing classic REC methods. HieA2G with the ResNet101 backbone even outperforms GSVA-7B \cite{xia2024gsva} based on MLLM. The promising results can be attributed to the hierarchical multi-modal semantic alignment design, which promotes a comprehensive understanding of information at different granularities.

\noindent \textbf{Results on Phrase Grounding.}
The main results on the Flickr30K Entities are shown in Table \ref{phrase}. HieA2G with ResNet101 improves performance over the previous SOTA MDETR on both val and test splits, suggesting that our method effectively enhances the multi-modal interactions.

\noindent \textbf{Results on RES.}
As shown in Table \ref{RES}, HieA2G outperforms the previous SOTA method ReLA \cite{liu2023gres} with the same Swin-B backbone. It also shows competitive performance on RefCOCO and RefCOCO+ datasets to MLLM-based LISA-7B \cite{lai2024lisa} and GSVA-7B. The results demonstrate that the comprehensive multi-modal representation ability of our HieA2G can contribute a lot to accurate segmentation for referring objects.

\noindent \textbf{Results on GRES.} GRES aims to generate masks for an arbitrary number of target objects. Unlike the previous SOTA GRES method ReLA using a simple binary classification branch for object-existence judgment, HieA2G has an explicit object-counting ability to facilitate accurate object perception in the generalized scenario. In Table \ref{GRES-1}, HieA2G with the Swin-B backbone achieves clear performance improvements over ReLA across all three evaluation sets on different metrics. It is even slightly better than the strong MLLM-based GSVA-7B in CIoU and GIoU. Besides generating high-quality masks, HieA2G with either backbone demonstrates outstanding performance in N-acc. and T-acc., indicating its robust object perception ability.

\subsection{Ablation Study}

\noindent \textbf{Effect of Hierarchical Multi-modal Semantic Alignment.} From the first to fourth rows of Table \ref{aba}, we perform an in-depth study of the HMSA module to validate its effectiveness. 
In the first row, removing all three-level alignments of the HMSA module leads to a significant decrease of 2.6\% in Pr@($\mathrm{F_1}$=1, IoU $\geq$ 0.5) and 5.4\% in N-acc.
Furthermore, we can find that removing any one of the alignments, including W2O, P2O, and T2I, all leads to performance degradation compared with the overall model in the seventh row. This demonstrates that different levels of alignment can refine the object embeddings of the object queries, further facilitating accurate grounding by combining them.

\noindent \textbf{Effect of Adaptive Grounding Counter.} In the last three rows of Table \ref{aba}, we test the effectiveness of AGC. The overall AGC is removed in the fifth row and replaced by the default threshold-based strategy \cite{he2023grec} to filter the output objects. We can observe a 13.9\% and 12.3\% performance drop in terms of Pr@($\mathrm{F_1}$=1, IoU $\geq$ 0.5) and N-acc., which reflects that our advanced adaptive selection strategy contributes a lot to the output of target objects. Then, we add the classifier in the sixth row, which achieves 12.6\% and 9.3\% performance gain in both metrics respectively. When combined with $\mathcal{L}_{con}$ in the last row, further improvement can be brought for all metrics.  This suggests that $\mathcal{L}_{con}$ is helpful to enhance the model’s object counting ability.

\subsection{Qualitative Analysis}
We visualize some qualitative examples of our method on the validation split of gRefCOCO dataset to discuss the strengths and weaknesses of HieA2G as shown in Figure \ref{fig:fig_case}.

\noindent \textbf{Analysis of Success Cases.}
As shown in (A) of Figure \ref{fig:fig_case}, our model can deal with various complex multi-target expressions in (a)-(d) and no-target expressions in (e)-(f). For example, HieA2G can count accurately ``\textit{two airplanes}'' in (a) with shared attributes, and can differentiate an ordinal number like ``2nd'' to detect the correct bus in (b). It can also explicitly recognize all specified target objects for complex text expressions in (b), (c), and (d). Moreover, HieA2G can grasp the fine-grained attribute details to reject the no-target expression ``\textit{the board in man hand}'' in (e). It has a comprehensive understanding of the global contextual information of all objects in (f), thereby rejecting to give a detection result due to no object in the image satisfying the description.

\noindent \textbf{Analysis of Failure Cases.} We show some failure cases of HieA2G in (B) of Figure \ref{fig:fig_case}. There are two main types of failure cases. \ding{182} For the three cases (a)-(c) in the first row, due to the ambiguous visual clues in the image,  HieA2G struggles to detect all target objects for the first two cases and fails to reject giving a target for the last case.
\ding{183} For the three cases (d)-(f) in the second row,  due to the occlusion of the key visual clues,  HieA2G fails to detect a desired object for the first two cases and gives a false negative target for the last no-target case. The analysis of failure cases reveals the limitations of HieA2G, while also shedding light on potential directions for our future research.

\section{Conclusion}

We propose a Hierarchical Alignment-enhanced Adaptive Grounding Network (HieA2G) for the challenging GREC task. The proposed Hierarchical Multi-modal Semantic Alignment (HMSA) module enables multi-level cross-modal interactions to achieve comprehensive and robust multi-modal understanding for better grounding. Adaptive Grounding Counter (AGC) determines the number of output targets dynamically to help select the outputs, effectively tackling the varying number of target objects in flexible referring expressions. The experimental results demonstrate the remarkable superiority and generalizability of the proposed HieA2G on multiple visual grounding tasks including REC, GREC, phrase grounding, RES, and GRES.

\bigskip

\bibliography{aaai25}

\end{document}